%% file: 00_main_paper.tex
\documentclass{article}

\usepackage{microtype}
\usepackage{graphicx}
\usepackage{subfigure}
\usepackage{amsfonts}
\usepackage{booktabs} 
\usepackage{multirow}

\usepackage{hyperref}
\usepackage{subcaption}
\usepackage{amsmath}

\usepackage{setspace}

\usepackage{amssymb,amsmath,amsthm,enumitem}

\newcommand\revise[1]{\textcolor{black}{#1}}

\usepackage[accepted]{mlsys2025}

\mlsystitlerunning{SparseTransX: Efficient Training of Translation-Based Knowledge Graph Embeddings Using Sparse Matrix Operations}

\usepackage{graphicx}
\usepackage{hyperref}
\usepackage{eso-pic}
\usepackage{xparse}

\begin{document}

\twocolumn[
\mlsystitle{SparseTransX: Efficient Training of Translation-Based Knowledge Graph Embeddings Using Sparse Matrix Operations}


\begin{mlsysauthorlist}
\mlsysauthor{Md Saidul Hoque Anik}{iu}
\mlsysauthor{Ariful Azad}{iu}
\end{mlsysauthorlist}


\mlsysaffiliation{iu}{Department of Computer Science and Engineering, Texas A\&M University, USA}

\mlsyscorrespondingauthor{Md Saidul Hoque Anik}{anik@tamu.edu}

\mlsyskeywords{Knowledge Graph, Sparse Linear Algebra, High-Performance Computing, MLSys}

\vskip 0.3in
\input{0_section_abstract}
]

\printAffiliationsAndNotice{}

\input{1_section_introduction}
\input{2_section_background}

\input{3_section_related_works}
\input{4_section_methodology}
\input{5_section_experimental_setting}
\input{6_section_results}
\input{7_section_discussion}

\input{8_section_conclusion}

\section*{Acknowledgements}
\revise{This research is supported by the NSF OAC-2112606 and OAC-2339607 grants and DOE DE-SC0022098 and DE-SC0023349 awards. 
}
 
\bibliography{references}
\bibliographystyle{mlsys2025}

\input{9_appendix}


\end{document}

%% file: 0_section_abstract.tex
\begin{abstract}
Knowledge graph (KG) learning offers a powerful framework for generating new knowledge and making inferences. Training KG embedding can take a significantly long time, especially for larger datasets. Our analysis shows that the gradient computation of embedding is one of the dominant functions in the translation-based KG embedding training loop. We address this issue by replacing the core embedding computation with SpMM (Sparse-Dense Matrix Multiplication) kernels. This allows us to unify multiple scatter (and gather) operations as a single operation, reducing training time and memory usage. We create a general framework for training KG models using sparse kernels and implement four models, namely TransE, TransR, TransH, and TorusE. Our sparse implementations exhibit up to 5.3x speedup on the CPU and up to 4.2x speedup on the GPU with a significantly low GPU memory footprint. The speedups are consistent across large and small datasets for a given model. Our proposed sparse approach can be extended to accelerate other \revise{translation-based (such as TransC, TransM, etc.) and non-translational (such as DistMult, ComplEx, RotatE, etc.) models as well. An implementation of the SpTransX framework is publicly available as a Python package in \href{https://github.com/HipGraph/SpTransX}{https://github.com/HipGraph/SpTransX}.}
\end{abstract}

%% file: 1_section_introduction.tex
\section{Introduction}
\label{introductoin}

Knowledge Graphs (KGs) are structured as directed graphs containing entities as nodes and relations as edges.
Each edge in a KG is typically stored as a triplet (head, relation, tail)—abbreviated as (h, r, t)—where head and tail are entities connected by a relation that denotes the nature of their interaction. Knowledge graph embedding (KGE) techniques map these entities and relations into a continuous vector space, enabling efficient computation and manipulation while preserving the underlying structural properties of the KG.
The entity and relation embeddings are widely used in many downstream tasks, such as KG
completion~\cite{TransE, chen2020knowledge}, entity classification~\cite{nickel2012factorizing}, and entity resolution~\cite{bordes2014semantic}.

Translational models~\cite{TransE, transR} are a widely used and effective class of KGE methods. 
These models represent entities and relations in a continuous vector space, where relations are interpreted as translations applied to entity embeddings. 
However, training translational KGE models for large-scale KGs is computationally intensive and incurs high memory overhead, especially when large batches are used. 
These challenges are due in part to the fact that current KGE implementations represent triplets as dense matrices and rely heavily on fine-grained scatter-gather computations during training.
This fine-grained computational model contributes to the following bottlenecks: (1) irregular memory access patterns from fine-grained operations on KGs and embeddings, which increase memory access costs, (2) increased backpropagation expenses due to more granular gradient computations, and (3) significant memory demands for dense matrices.
In this paper, we address these issues by proposing a sparse-matrix representation of the KG and utilizing highly optimized sparse-matrix operations to streamline KGE training, thereby reducing both computational and memory bottlenecks.

Expressing graph operations through sparse linear algebra has been highly effective for developing efficient and scalable graph neural networks (GNNs). As a result, popular graph machine learning libraries, such as PyTorch Geometric (PyG) \cite{fey2019fast} and DGL \cite{wang2019deep}, utilize optimized implementations of sparse-dense matrix multiplication (SpMM).
Despite the widespread success of sparse operations in GNNs, existing KGE libraries have yet to adopt sparse operations for training KGE models. Even models utilizing sparse embeddings, such as TranSparse~\cite{transSparse}, store embeddings as dense matrices, limiting their ability to fully leverage sparse matrix operations.

\begin{table}[h]
\caption{200 epoch training time breakdown of a TransE model using the sparse approach compared to a non-sparse approach. The time shown is the average training time taken for 7 datasets listed in Table \ref{table:datasets}. The GPU is a single NVIDIA A100-SXM4 with 40 GB VRAM. The CPU is an AMD EPYC 7763 (Milan) CPU with 64 cores and 512GB DDR4 memory.}
\centering
\label{intro-table}
\begin{tabular}{|cc|c|c|}
\hline
\multicolumn{2}{|c|}{}                                                  & \textbf{Sparse} & \textbf{Non-Sparse } \\ 
& & & \textbf{(TorchKGE)} \\
\hline
\multicolumn{1}{|c|}{\multirow{3}{*}{\textbf{CPU}}} & \textbf{Forward}  & 74.86           & 299.2               \\ \cline{2-4} 
\multicolumn{1}{|c|}{}                              & \textbf{Backward} & 166.59          & 919.17              \\ \cline{2-4} 
\multicolumn{1}{|c|}{}                              & \textbf{Step}     & 15.4            & 15.95               \\ \hline
\multicolumn{1}{|c|}{\multirow{3}{*}{\textbf{GPU}}} & \textbf{Forward}  & 18.2            & 48.8                \\ \cline{2-4} 
\multicolumn{1}{|c|}{}                              & \textbf{Backward} & 17.49           & 89.51               \\ \cline{2-4} 
\multicolumn{1}{|c|}{}                              & \textbf{Step}     & 0.4             & 0.45                \\ \hline
\end{tabular}

\end{table}

One of this paper's main contributions is the development of sparse formulations for several popular translation-based KGE models. 
Adapting different translation models to sparse operations presents unique challenges, as each model interprets translations differently. 
For instance, TransE~\cite{TransE} uses a single embedding space for both entities and relations, while TransR~\cite{transR} uses separate spaces. Despite these differences, we designed a unified framework that allows diverse translation models to be represented through sparse matrices and mapped to sparse matrix operations like SpMM. We collectively refer to these sparse variants of translation-based embedding methods as SpTransX.

We develop a comprehensive library based on our sparse formulation. This library consolidates most computations into several SpMM function calls, allowing optimized SpMM to directly accelerate the overall runtime of KGE training. \revise{We also discuss how to extend this concept to other non-translational models such as DistMult~\cite{DistMult} or ComplEx~\cite{ComplEx} in Appendix \ref{A:non_trans}.} We observe that SpTransX models significantly outperform established knowledge graph frameworks, such as TorchKGE and DGL-KE, particularly in terms of training time and GPU memory usage. For example, the average improvement in training time for the TransE model is illustrated in Table \ref{intro-table}.

Overall, this paper presents the following contributions:
\vspace{-0.2cm}
\begin{enumerate}
    \item {\bf Sparse Formulations of Translation-Based KGE Models:} We introduce sparse formulations for translation-based KGE models, enabling the mapping of KGE computations to SpMM and leveraging well-established SpMM techniques in model training.
    \item {\bf Development of an Optimized Library:} Our library incorporates various optimization techniques, including SIMD vectorization, loop unrolling, cache blocking, tiling, and WARP-level GPU optimization, to enhance performance. As a result, SpTransX models significantly outperform established knowledge graph frameworks, such as TorchKGE and DGL-KE.
    \item {\bf Enhanced Large-Batch Training:} By reducing memory requirements, SpTransX facilitates large-batch training on memory-limited GPUs.

\end{enumerate}




%% file: 2_section_background.tex
\section{Background}
Knowledge graph training is performed by learning the representations or embeddings of the entities and their corresponding relations on a set of training triplets or subgraphs. Each triplet or edge (in subgraph) contains a valid combination of subject (head), predicate (relation), and object (tail). Once trained, the embeddings can illustrate their semantic meaning and structure, enabling them to effectively perform reasoning-based tasks such as link prediction and entity classification. The training typically uses machine learning techniques and involves a gradient descent algorithm. The exact forward propagation process can vary depending on the model type. Translation-based models, such as TransE, TransR, etc, are widely used due to their simple yet effective way of capturing relations between entities. The training can also be done by using bilinear methods (DistMult \cite{yang2014embedding}, RESCAL \cite{nickel2011three}), deep learning and convolution (ConvKB \cite{nguyen2017novel}), or Graph Neural Networks (R-GCN \cite{schlichtkrull2018modeling}).

Training a translation-based knowledge graph embedding typically involves taking a list of triplets (head, tail, relation index) and optimizing their corresponding embeddings to minimize the distance between the $\Vec{tail}$ and $\Vec{head} + \Vec{relation}$. The translational models vary based on (1) the linear transformation applied to the entities and relations and (2) the distance metric. The linear transformation can be applied to (a) individual entities/relations, (b) head - tail, or (c) the overall head - tail + relation. The measurement can be in a typical Euclidean space (L1 or L2) or a toroidal (wraparound) space distance (L1 torus or L2 torus) function. The training is typically done in batches, where a `batch' of head, tail, and relations are fetched for training instead of single ones.

\begin{figure}[h]
\centering     
\subfigure[Forward]{\label{fig:gather}\includegraphics[width=35mm]{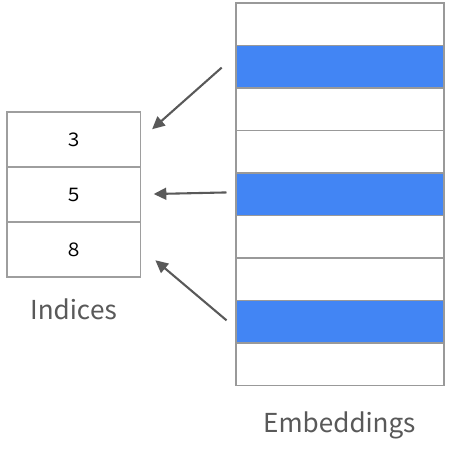}}
\hspace{5pt}
\subfigure[Backward]{\label{fig:scatter}\includegraphics[width=35mm]{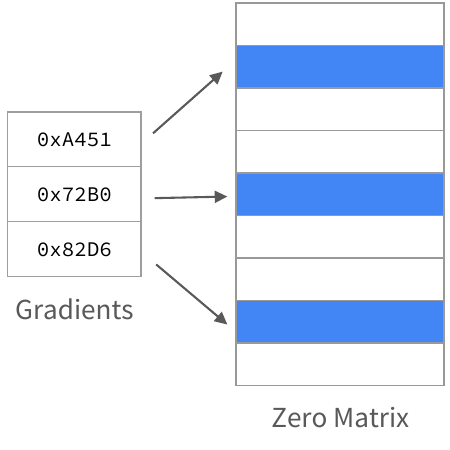}}
\caption{Scatter and Gather operation in translational KG training}
\end{figure}

\begin{figure*}[!t]
\centering
\includegraphics[width=0.90\textwidth]{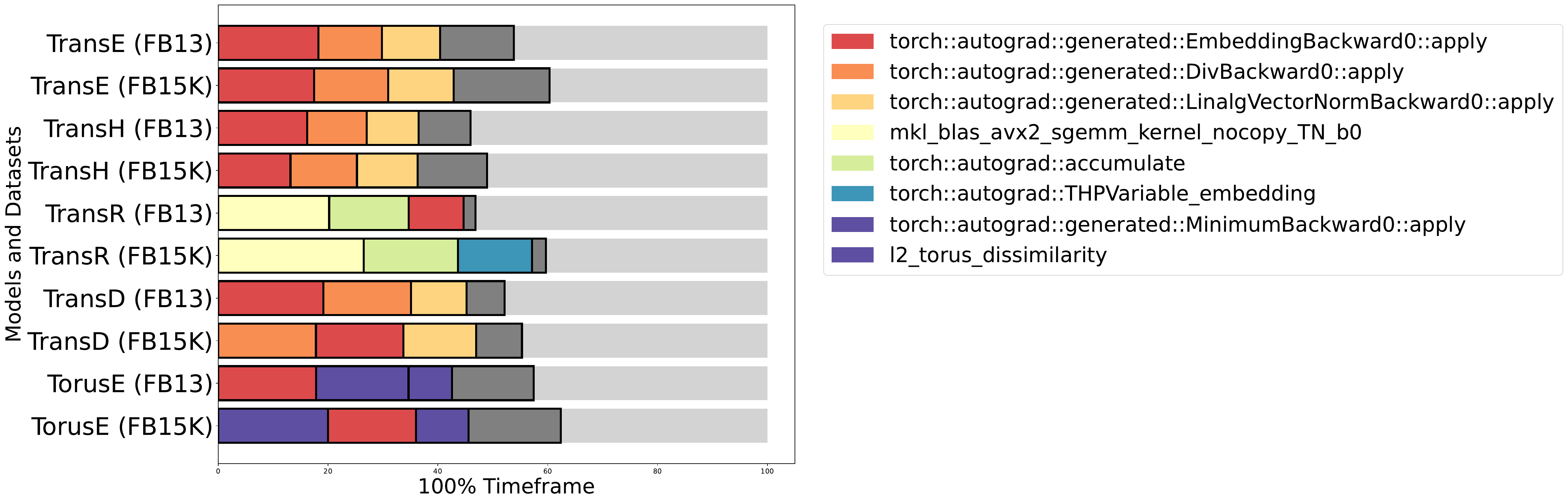}
\caption{Top three CPU intensive functions for various translation-based KGE models and datasets (indicated in brackets). The redness represents the popularity of a function among models. The dark red box indicates that the corresponding function is used in several different models. \revise{Blue/Purple} indicates that the function is typically exclusive to the current model. The dark gray box indicates the dataset loading time. The light gray box indicates the rest of the training time.}
\label{fig:bottlenecks}
\end{figure*}

The training process starts with triplets with the index position of the head, tail, and relation entities. In each epoch, embeddings are fetched from the indices, and linear transformation is applied to them to compute the final loss. This means the forward propagation involves several (typically three or more) `gather' operations (see Figure \ref{fig:gather}) that collect the index batch's head, tail, and relation embeddings. Some models also require one or more transform matrices (may be based on the relation), which are also gathered in sthis step. Consequently, the backward propagation performs the opposite, the `scatter' operations that distribute gradients across the corresponding indices (see Figure \ref{fig:scatter}) 

These individual operations, especially the gradient computations in the backward step, can take up around 40\% of the CPU's training time (see Figure \ref{fig:bottlenecks}). In particular, we observe that embedding gradient computation is among the top three CPU-intensive functions for most translational models.

%% file: 3_section_related_works.tex
\section{Related Work}
\subsection{Translational Models for KGE}
Translation-based models represent entities and relations in a continuous vector space, interpreting relations as translations operating on entity embeddings. Several well-known models follow this approach, including TransE~\cite{TransE}, TransR~\cite{transR}, TransH~\cite{TransH}, TransD~\cite{transD}, TransA~\cite{transA}, TransG~\cite{transG},
TransC~\cite{lv2018differentiating}, TransM~\cite{fan2014transition}, 
TorusE~\cite{torusE}, and KG2E~\cite{KG2E}. Each model varies in how it represents the head, relation, and tail embeddings to capture relational semantics effectively.
For instance, TransE embeds entities and relations in the same vector space $\mathbb{R}^d$, assuming that relations can be modeled as a simple addition between the head and tail entities. In contrast, TransR utilizes distinct vector spaces for entities and relations, allowing it to better capture heterogeneous relation types, while TransE struggles with symmetric and one-to-many relations.
Some models, like TransH, introduce translations on hyperplanes to address the limitations of basic Euclidean embeddings. More recently, models such as rotatE~\cite{rotatE} have enabled translations within hyperbolic space instead of Euclidean space, allowing for better representation of hierarchical structures commonly found in some knowledge graphs.
It has been observed that translation-based models are typically more computationally efficient compared to semantic matching models that use a bilinear score function, such as DistMult \cite{DistMult}, RESCAL \cite{RESCAL}, and ComplEx \cite{ComplEx}. This efficiency, along with their adaptability across different KG structures, makes translation-based models a popular choice for large-scale knowledge graph applications.

\subsection{KGE frameworks}
Several frameworks are available for training knowledge graphs (KGs). Some, like TorchKGE \cite{boschin2020torchkge} and DGL-KE \cite{zheng2020dgl}, are specifically designed for this purpose. Others, such as PyTorch Geometric \cite{fey2019fast} and GraphStorm \cite{zheng2024graphstorm}, offer facilities for training KG models in addition to modules for training graph neural networks. 

Many frameworks are built on top of the PyTorch Framework, including TorchKGE, PyKeen \cite{ali2021pykeen}, PyTorch Geometric, etc. AmpliGraph \cite{ampligraph} has Tensorflow 2.0 backend. Some frameworks support hybrid backends, such as DGL-KE or OpenKE \cite{han2018openke}. DGL-KE supports PyTorch and MXNet as the backend. OpenKE supports PyTorch, Tensorflow, and C++ as the backend. Most frameworks have support for Python. 

Some frameworks, such as Pykg2vec \cite{yu2019pykg2vec} or DGL-KE, choose not to use the autograd feature of the backend ML, such as PyTorch, and implement their custom gradient update mechanism. PyKeen is designed to be highly extensible and uses a modular code base. It features automatic memory optimization support that generates sub-batches when the user-defined batch does not fit in the memory.

Most frameworks, such as TorchKGE, PyG, and PyKeen, use PyTorch's embedding module directly to store entity and relation embeddings.  Others, such as DGL-KE, convert the training triplets into DGL graphs before training. DGL-KE, PyTorch BigGraph \cite{lerer2019pytorch}, PyKeen, and several other frameworks allow multi-CPU and multi-GPU training using Python and distributed frameworks such as DGL or PyTorch Lightning.

\subsection{Sparse Operations in Graph ML} 
Expressing graph operations through sparse linear algebra has proven highly effective for developing efficient and scalable graph learning algorithms. 
For example, the forward and backward propagation in graph convolutional networks (GCNs) and graph attention networks (GATs) can be optimized with sampled dense-dense matrix multiplication (SDDMM), sparse-dense matrix multiplication (SpMM), or their combination, known as FusedMM~\cite{fey2019fast, wang2019deep, rahman2021fusedmm}. 
Similarly, SpMM and sparse-sparse matrix multiplication (SpGEMM) are widely used in algorithms for graph embedding~\cite{ranawaka2024distributed, rahman2020force2vec}, clustering~\cite{azad2018hipmcl}, and visualization~\cite{rahman2020batchlayout}. 
Consequently, popular graph machine learning libraries, such as PyTorch Geometric (PyG) \cite{fey2019fast} and DGL \cite{wang2019deep}, rely on optimized implementations of SpMM, SpGEMM, and SDDMM available in vendor-provided libraries like cuSparse, MKL, or open-source libraries such as iSpLib~\cite{hoque2024isplib}, FeatGraph \cite{hu2020featgraph}, and SparseTIR \cite{ye2023sparsetir}.
Despite the wide success of sparse operations in GNNs and graph embeddings, to the best of our knowledge, existing knowledge graph embedding libraries do not leverage sparse operations for training KGE models.


%% file: 4_section_methodology.tex
\section{Methodology}
\subsection{Sparse Approach}
We observe that the embedding extraction operation and its gradient computation is a bottleneck in the training of many translational models (Figure \ref{fig:bottlenecks}). We tackle this by replacing the typical embedding extraction process with Sparse-Dense Matrix Multiplication (SpMM). We form a sparse incidence matrix out of the training triplets so that multiplying it with the embedding matrix would directly generate at least a portion of the scores for each triplet.

The sparse approach unifies the embedding gather operations for entities in forward propagation and scatter operations for gradients in backward propagation.
This unified framework enables us to leverage high-performance matrix multiplication techniques, such as loop unrolling, cache blocking, tiling, and WARP-level GPU primitives. Additionally, we can apply advanced parallelization methods, including dynamic load balancing across threads and code generation, as well as more efficient SIMD (Single Instruction, Multiple Data) vectorization.

In the following subsections, we briefly discuss how to perform training in a sparse approach using SpMM instead of regular embedding extraction for several translation-based models. \revise{Both forward and backward propagation of our approach benefit from the efficiency of a high-performance SpMM (proof shown in Appendix \ref{A:backprop}). This concept also extends broadly to various other knowledge graph embedding (KGE) methods as well, including DistMult, ComplEx, and RotateE (detailed formulations are provided in Appendix \ref{A:non_trans}). The sparsity of our formulation and related computational complexity are discussed in Appendix \ref{A:density} and \ref{A:complexity}.}

\subsection{Adjacency Matrix Formulation}
We analyze the score function of several translation-based models and observe that many models such as TransE, TransR, and TransH take head, tail, and relation - and compute either (a) (head - tail) or (b) (head - tail + relation) expression before applying additional linear projections as needed. 
For simplicity, we refer to these as `ht' and `hrt' expressions, respectively.
 Table \ref{table:common-models} lists a few of such models and their corresponding score functions. For some models, the expressions mentioned earlier are apparent, and for others, we need to perform minor algebraic rearrangements. These formulations are listed from subsection \ref{transe_formulation} to \ref{toruse_formulation}.
\vskip -.1in

\begin{table}[h]
\centering
\caption{Translational models with common expressions in score function $f_r(h,t)$}
\label{table:common-models}
\begin{center}
\begin{small}
\begin{sc}
\begin{tabular}{cc}
\toprule
\textbf{Model}      & \textbf{Scoring Function} \\
\midrule
TransE \cite{TransE} & $||\Vec{h} + \Vec{r} - \Vec{t}||$                 \\
TransH \cite{TransH}   & $||\Vec{h_\perp} + \Vec{d_r} - \Vec{t_\perp}||$                        \\
TransR \cite{transR}    & $||M_r\Vec{h} + \Vec{r} - M_r\Vec{t}||$                         \\
TorusE \cite{torusE}  & $||\Vec{h} + \Vec{r} - \Vec{t}||$                   \\
TransA \cite{transA}   & \tiny{$|\Vec{h} + \Vec{r} - \Vec{t}|^T W_r |\Vec{h} + \Vec{r} - \Vec{t}|$}                        \\
TransC \cite{lv2018differentiating}     & $||\Vec{h} + \Vec{r} - \Vec{t}||^2_2$                         \\
TransM \cite{fan2014transition} & $w_r||\Vec{h} + \Vec{r} - \Vec{t}||$                                          \\                        
\bottomrule
\end{tabular}
\end{sc}
\end{small}
\end{center}
\vskip -0.1in
\end{table}

\vskip -.1in
Instead of gathering head, tail, and relations individually from the indices and then computing the {\em ht} and {\em hrt} expressions, we can directly get this result by forming an incidence matrix. The following subsection describes how we can compute {\em ht} and {\em hrt} expressions using sparse-dense matrix multiplication.

\begin{figure*}[t]
\centering     
\subfigure[$(\textit{h}-\textit{t})$ Computation. All cells in the highlighted row are zero except h-idx and t-idx. For the highlighted row, h-idx = 5, t-idx = 15, entity-count = 22.]
{\label{fig:ht}\includegraphics[width=0.9\columnwidth]{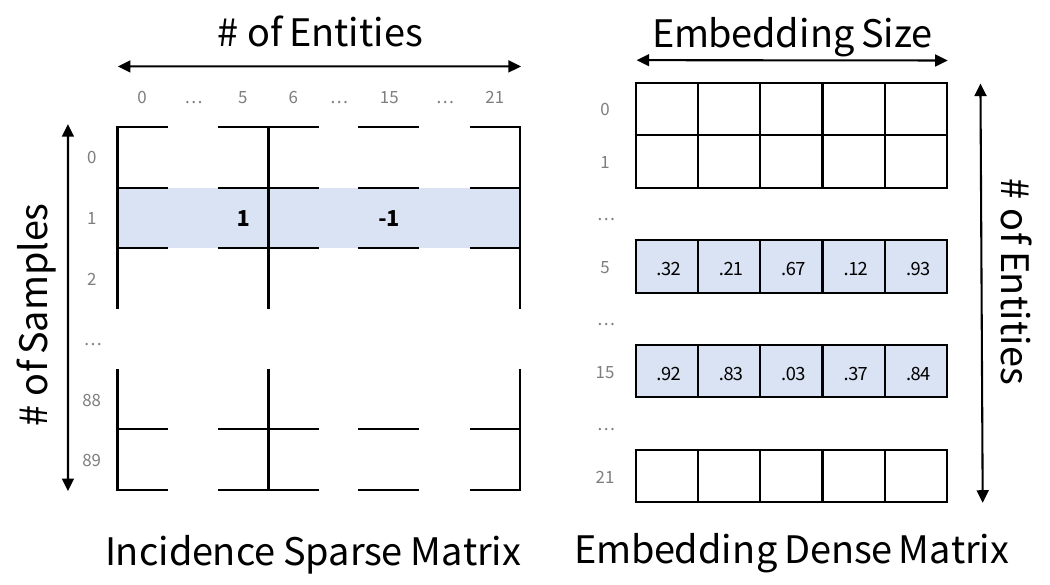}}
\hspace{10pt}
\subfigure[$(\textit{h}+\textit{r}-\textit{t})$ Computation. All cells in the highlighted row are zero except h-idx, t-idx, and r-idx + entity-count. For the highlighted row, h-idx = 5, t-idx = 15, entity-count = 20, r-idx = 2.]{\label{fig:hrt}\includegraphics[width=1.1\columnwidth]{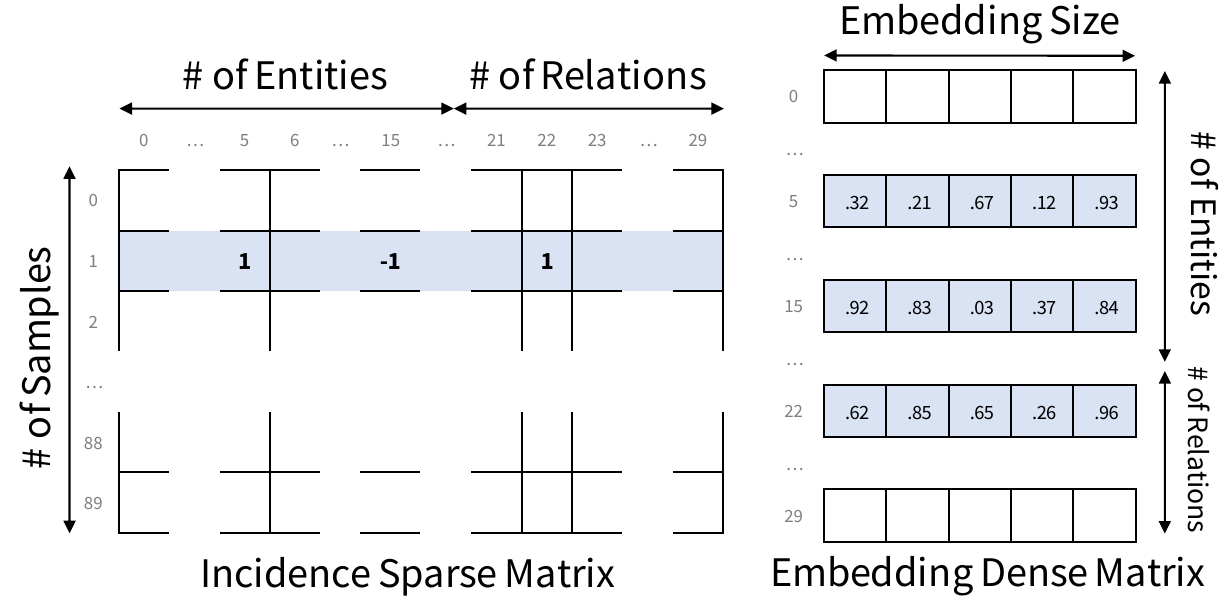}}

\caption{Computing common expressions using SpMM. Only the highlighted row is populated for demonstration.}
\label{fig:htr_ht_computation}
\end{figure*}

\subsubsection{ht or (head - tail) computation}
\label{sec:ht}
Let the knowledge graph contain 
$N$ entities and $M$ triples in the training data, with an embedding size (dimension) denoted by $d$.
To compute the \emph{ht} expression, we store entity embedding in a dense matrix $\mathbf{E} \in \mathbb{R}^{N\times d}$, where each row stores the embedding of an entity.
We store the training triples in a sparse incidence matrix $\mathbf{A} \in \{-1,0,1\}^{M\times N}$, where the rows represent the training triplets and the columns represent entities. 
For a triplet, the corresponding column of a head or tail index is filled with the coefficient of the head or tail. In the expression head - tail, the coefficient of the head is $+1$ and $-1$ for the tail.  
This implies that each row of the incidence matrix contains exactly two nonzero entries.
Once we multiply this incident sparse matrix $\mathbf{A}$ with the embedding matrix $\mathbf{e}$, we get the array of (head - tail) for the corresponding training triplets. Figure \ref{fig:ht} shows an example of this calculation. This computed expression can be used to complete the score calculation.

\subsubsection{hrt or (head + relation - tail) computation}
\label{sec:hrt}
Evaluating this expression requires accessing two separate dense matrices when entity and relation embeddings are stored individually. We can still compute this expression in a single sparse-dense matrix multiplication if we stack the entity and relations horizontally in the incidence sparse matrix and vertically as an embedding dense matrix.

Let the knowledge graph contain 
$R$ relations.
To compute the \emph{hrt} expression with a single SpMM operation, we store the entity and relation embeddings in the same dense matrix $\mathbf{E} \in \mathbb{R}^{(N+R)\times d}$, where the first $N$ row stores the embeddings of entities and the last $R$ rows store the embeddings of relations.
For this computation, we store the training triples in a sparse incidence matrix $\mathbf{A} \in \{-1,0,1\}^{M\times (N+R)}$, where the rows represent the training triplets and the columns represent entities and relations.
As before, we place the expressions' coefficients in the corresponding columns ($+1$ for head and relation, $-1$ for tail).  The relation associated with each triple is represented by placing a $+1$ in the corresponding column for that relation.  
Note that we offset the relation index by the total number of entities in the incidence matrix $\mathbf{A}$. This ensures that, when multiplied, the relation index aligns correctly with the corresponding relation embedding located just below the entity embeddings.
Finally, we multiply the sparse matrix with the combined dense embedding matrix to get the hrt expression result. Figure \ref{fig:hrt} shows an example of this computation.

The following subsections contain the implementation of four translational models using the sparse approach. Throughout the rest of the paper, we refer to these four implementations collectively as \textbf{SparseTransX}, or \textbf{SpTransX} in short.

\subsection{TransE Formulation}
\label{transe_formulation}
For triplets ($\Vec{h}$, $\Vec{r}$, $\Vec{t}$), where $\Vec{h}$ is the head entity vector, $\Vec{r}$ is the relation entity vector, and $\Vec{t}$ is the tail entity vector, TransE tries to enforce the following for a training set U:
\begin{spacing}{0.2}
\begin{equation} \label{eq1}
\begin{split}
    \forall ( \Vec{h}, \Vec{t}, \Vec{r}) \in U, & \\
     & \Vec{h} + \Vec{r} \approx \Vec{t} \\
     \implies & \Vec{h} + \Vec{r} - \Vec{t}\approx \Vec{0} \\ \\
    \end{split}
\end{equation}
\end{spacing}
For TransE, a normalization function (L1 or L2) is typically applied to this expression to get the final score. We can directly obtain this expression using the hrt computation method discussed in subsection \ref{sec:hrt}.

\subsection{TransR Formulation}
\label{transr_formulation}
The TransR model applies a linear projection to the head and tail before computing the score. For a projection matrix $M_r$ corresponding to relation $\Vec{r}$, TransR tries to enforce the following translation:

\begin{spacing}{0.2}
\begin{equation} \label{eq2}
\begin{split}
    \forall ( \Vec{h}, \Vec{t}, \Vec{r}) \in U, & \\
     & M_r\Vec{h} + \Vec{r} \approx M_r\Vec{t} \\
     \implies & M_r\Vec{h} + \Vec{r} - M_r\Vec{t}\approx \Vec{0} \\
     \implies & M_r(\Vec{h} - \Vec{t})+ \Vec{r}\approx \Vec{0} \\
     \\
    \end{split}
\end{equation}
\end{spacing}

After rearrangement, we see that it contains the (head-tail) expression. This can be computed using the ht computation method discussed in subsection \ref{sec:ht}.

\subsection{TransH Formulation}
\label{transh_formulation}
TransH (Translating Embeddings on Hyperplanes) extends TransE by allowing each relation to have its hyperplane, addressing the limitation that a single translation vector cannot handle 1-to-N, N-to-1, and N-to-N relations effectively. In TransH, each relation $\Vec{r}$ is associated with a hyperplane characterized by a normal vector $\Vec{w_r}$ and a translation vector $\Vec{d_r}$. The projection of entities onto the hyperplane is then used in the translation. It tries to enforce the following:

\begin{spacing}{0.2}
\begin{equation} \label{eq3}
\begin{split}
    \forall ( \Vec{h}, \Vec{t}, \Vec{r}) \in U, & \\
     & \Vec{h_{\perp}} + \Vec{d_r} \approx \Vec{t_{\perp}} \\
    \end{split}
\end{equation}
\end{spacing}

Where, 
\begin{spacing}{0.5}
$$\Vec{h_{\perp}} = \Vec{h} - (\Vec{w_r}^T\cdot \Vec{h})\Vec{w_r}$$
$$\Vec{t_{\perp}} = \Vec{t} - (\Vec{w_r}^T\cdot \Vec{t})\Vec{w_r}$$
\end{spacing}

Substituting these values in Equation \ref{eq3}, we find that for every triplet, TransH is trying to enforce:

\begin{spacing}{0.5}
\begin{equation*}
\begin{split}
    & \Vec{h} - (\Vec{w_r}^T\cdot \Vec{h})\Vec{w_r} + \Vec{d_r} \approx \Vec{t} - (\Vec{w_r}^T\cdot \Vec{t})\Vec{w_r}\\
    \implies & \Vec{h} - (\Vec{w_r}^T\cdot \Vec{h})\Vec{w_r} + \Vec{d_r} - \Vec{t} + (\Vec{w_r}^T\cdot \Vec{t})\Vec{w_r} \approx \Vec{0} \\
    \implies & \Vec{h} - \Vec{t} + \Vec{d_r} - (\Vec{w_r}^T\cdot \Vec{h})\Vec{w_r} + (\Vec{w_r}^T\cdot \Vec{t})\Vec{w_r} \approx \Vec{0} \\
    \implies & (\Vec{h} - \Vec{t}) + \Vec{d_r} - \Vec{w_r}^T\cdot(\Vec{h} - \Vec{t})\Vec{w_r} \approx \Vec{0} \\
    \end{split}
\end{equation*}
\end{spacing}

We observe that the final arrangement contains two expressions of ht. This can be computed using the ht computation method discussed in subsection \ref{sec:ht}.

\subsection{TorusE Formulation}
\label{toruse_formulation}
The TorusE model is very similar to TransE regarding the score function. It typically uses L1/L2 torus distance instead of regular L1/L2 norm and only works with the fractional components of the embeddings.

Just like TransE, it also tries to enforce the following:
\begin{spacing}{0.5}
\begin{equation} \label{eq4}
\begin{split}
    \forall ( \Vec{h}, \Vec{t}, \Vec{r}) \in U, & \\
     & \Vec{h} + \Vec{r} \approx \Vec{t} \\
     \implies & \Vec{h} + \Vec{r} - \Vec{t}\approx \Vec{0} \\ \\
    \end{split}
\end{equation}
\end{spacing}

We can directly obtain this expression using the hrt computation method discussed in subsection \ref{sec:hrt}.


\begin{figure}[h]
    \centering
    \includegraphics[width=0.45\textwidth]{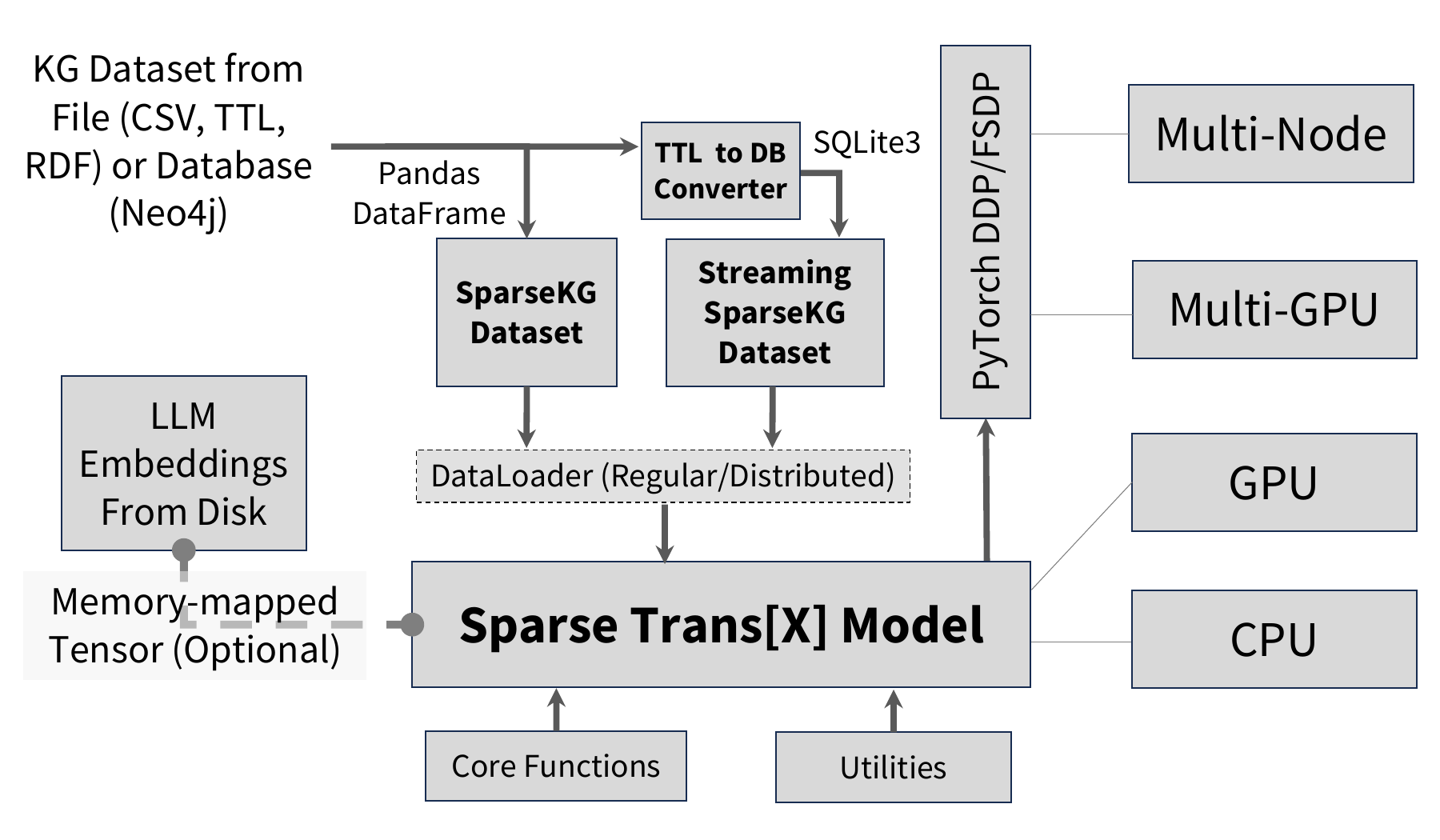}
    \caption{SparseTransX Framework}
    \label{fig:fastkg}
    \end{figure}
    
\subsection{SparseTransX Framework}
We develop a general framework for SpTransX model training to enable efficient translation-based model training for large KG datasets. The framework is implemented using PyTorch 2.3 and consists of \revise{several} modules, which are briefly described below.

\subsubsection{SparseTransX Models}
This module contains the sparse implementations of the translational models. These implementations are agnostic to the sparse matrix library used underneath. The models have built-in support for streaming embeddings from disc storage when the embeddings are too large to fit in CPU memory. This streaming model support is implemented using PyTorch memory-mapped tensors. Researchers often use Large Language Model (LLM) embeddings such as BERT \cite{devlin-etal-2019-bert}, T5 \cite{colin2020exploring}, or GPT \cite{radford2018improving} to perform knowledge graph completion \cite{wang2022simkgc, kim2020multi} and want to start with pre-trained embeddings that are typically too large to fit on CPU memory. Such training can be performed using this feature of the framework. Finally, this module also has functionalities for calculating scores, predicting links, and classifying entities in addition to the training loop.

\subsubsection{Dataloaders}
Our framework contains various dataloaders for shared and distributed training. It supports several standard knowledge graph formats, such as TTL, RDF, and CSV. Additionally, it contains a streaming dataset module for datasets that are too large to fit in memory. When invoked, it creates an SQLite representation of the knowledge graph and stores the entity-index mapping in the database along with the triplets. All dataloaders connect to the sparse model input using a common interface.

\subsubsection{Utilities and Core Functions}
This module consists of the sparse adjacency matrix builder described in subsection \ref{sec:ht} and \ref{sec:hrt}, an efficient sparse negative sampler, and the matrix multiplier interface.



%% file: 5_section_experimental_setting.tex
\section{Experimental Setting}
We implement the SpTransX models using PyTorch Framework and compare their total training time and GPU memory allocation with other well-known KG frameworks. We run these experiments for 7 datasets consisting of various sizes on a single CPU and a single GPU system separately. \revise{We provide a guideline to reproduce the experiment for one of the datasets in Appendix \ref{A:AE}. Although our framework supports distributed training using standard PyTorch data-parallel libraries, this paper focuses on sparse techniques and presents experiments conducted on a single GPU only. A preliminary result on distributed training of a large knowledge graph dataset, COVID-19 \cite{tabassum2024knowledge}, is shown in Appendix \ref{A:scaling}.}

\subsection{Datasets}
Below are the 7 datasets used in the experiments.
    \begin{table}[h]
\caption{Knowledge graph datasets}
\label{table:datasets}
\vskip 0.15in
\begin{center}
\begin{small}
\begin{sc}
\begin{tabular}{ccccc}
\toprule
  &     &    & Training \\
  Dataset      &   Entity          &   Relations          & Triplets \\
\midrule
FB15k    & 14951  & 1345  & 483142  \\
FB15k237 & 14541  & 237   & 272115  \\
WN18     & 40943  & 18    & 141442  \\
WN18RR   & 40943  & 11    & 86835  \\                    
FB13     & 67399  & 15342 & 316232  \\
YAGO3-10 & 123182 & 37    & 1079040 \\
BioKG    & 93773  & 51    & 4762678 \\    
\bottomrule
\end{tabular}
\end{sc}
\end{small}
\end{center}
\vskip -0.1in
\end{table}
\subsection{Frameworks and Models}
\label{fw-models}
For comparison, we pick three popular KG frameworks: TorchKGE, PyTorch Geometric, and DGL-KE. PyTorch Geometric (or PyG) supports the TransE model, while DGL-KE supports the TransE and TransR models. TorchKGE supports all four models: TransE, TransR, TransH, TorusE.
\subsection{Training Loop}
We prepare 11 separate scripts (SpTransE, SpTransR, SpTransH, SpTorusE, transe-torchkge, transr-torchkge, transh-torchkge, toruse-torchkge, transe-dglke, transr-dglke, and transe-pyg) to train the models on various datasets. Each script receives the dataset name as a command-line argument. The dataset is loaded from a shared repository. All frameworks use the same training configuration (learning rate: 0.0004, margin: 0.5), dissimilarity function (L2 or L2 torus), and loss function (MarginRankingLoss), and run for 200 epochs. Batch size and embedding dimensions are selected to maximize accuracy while utilizing available GPU memory (see subsection \ref{hyperparam_selection}). The following table lists the dimensions and batch sizes used for different models. 

    \begin{table}[h]
\caption{Training configuration for CPU and GPU. \revise{Reduced embedding and batch size for TransR and TransH due to memory limitation}.}
\label{table:trg-config}
\vskip 0.15in
\begin{center}
\begin{small}
\begin{sc}
\begin{tabular}{ccc}
\toprule
  Model  & Embedding       & Batch     \\
\midrule
TransE & 1024            & $12 \times 32768$ \\
TorusE & 1024            & $12 \times 32768$ \\
TransR & 128             & $2 \times 32768$  \\
TransH & Ent=128,Rel=128 & $32768$    \\    
\bottomrule
\end{tabular}
\end{sc}
\end{small}
\end{center}
\vskip -0.1in
\end{table}

The negative samples are generated once per positive sample and are pre-generated outside the training loop.

\subsection{System and Profiler Details}
All experiments are run on the NERSC Perlmutter system. The GPU experiments run on a single NVIDIA
A100-SXM4 GPU with 40 GB VRAM. The CPU experiments run on an AMD EPYC 7763 (Milan) CPU with 64 cores and 512GB DDR4 memory. 

We use Python's $time$ module to measure training time and its breakdown. PyTorch's CUDA module is used to measure peak memory usage in GPU experiments. Finally, Linux's $perf$ tool is used to measure the cache miss rate and FLOPs count for CPU experiments.

\subsection{SparseTransX Configuration}
In the SparseTransX framework, users can choose to use any high-performance SpMM to perform model training.
We select iSpLib SpMM \cite{hoque2024isplib} for CPU training and DGL g-SpMM \cite{wang2019deep} for GPU training. \revise{CSR (Compressed Sparse Row) format is used for iSpLib, and COO (Coordinate) format is used for DGL as per the library requirement. The framework automatically generates sparse minibatches in the correct format when the user specifies the underneath SpMM library.}

%% file: 6_section_results.tex
\section{Results}
    \subsection{Hyperparameter Selection}
    \label{hyperparam_selection}
\begin{figure}[h]
    \centering
    \includegraphics[width=0.45\textwidth]{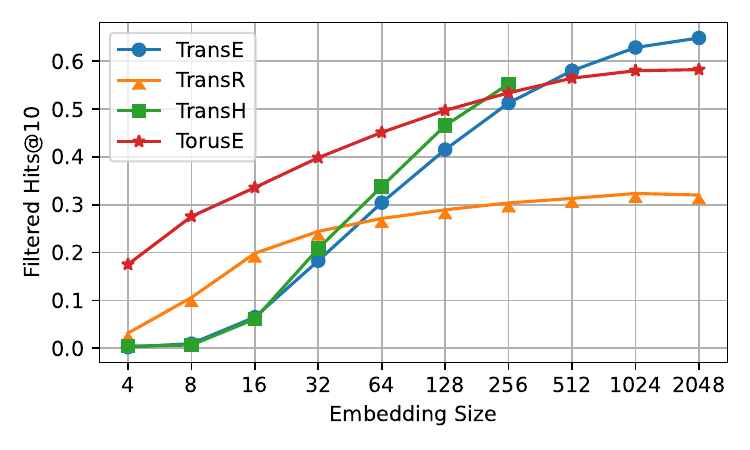}
    \caption{Hits@10 accuracy w.r.t. embedding size \revise{for FB15K dataset}. 100 epoch training with a batch size of 32768 and relation entity dimension as 8 (for TransH model). \revise{The TransH model encounters out-of-memory issues when the embedding size exceeds 256. Other models converge at an embedding size of approximately 2048 and show no improvement in Hits@10 accuracy for larger embeddings.}}
    \label{fig:acc-emb}
    \end{figure}

    Knowledge graphs are primarily used for link prediction tasks \cite{gregucci2023link}. Hits@10 is a popular measurement of link prediction accuracy. We train the KG models on various embedding sizes and plot the corresponding Hits@10 accuracy in Figure-\ref{fig:acc-emb}. We observe that accuracy increases as the entity embedding size grows. \revise{We keep the number of positive and negative edges equal within a batch for each model.}

     \begin{figure}[ht]
    \centering
    \includegraphics[width=0.45\textwidth]{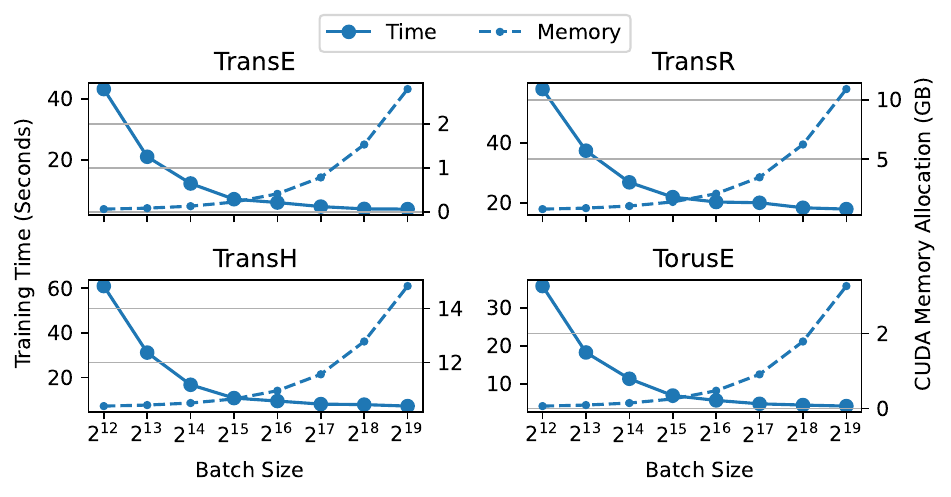}
    \caption{Training time and GPU memory allocation w.r.t. Batch Size. 100 epoch training with entity dimension as 128 and relation dimension as 8 (for TransH model).}
    \label{fig:batch-corr}
    \end{figure}

    Another significant hyperparameter is the batch size. We plot model training time and GPU memory allocation for various batch sizes in Figure \ref{fig:batch-corr}. We observe that maximum CUDA memory utilization is possible when the largest batch size is used. It also corresponds to the fastest training time.

    \subsection{Training Performance}
    We measure the total training time, GPU memory allocation, CPU Cache miss, and FLOPs count for various datasets on the available models of the frameworks mentioned in subsection \ref{fw-models}.

    \subsubsection{Training Time}
\begin{figure*}[btp]
\centering     
\subfigure[CPU]{\label{fig:a}\includegraphics[width=140mm]{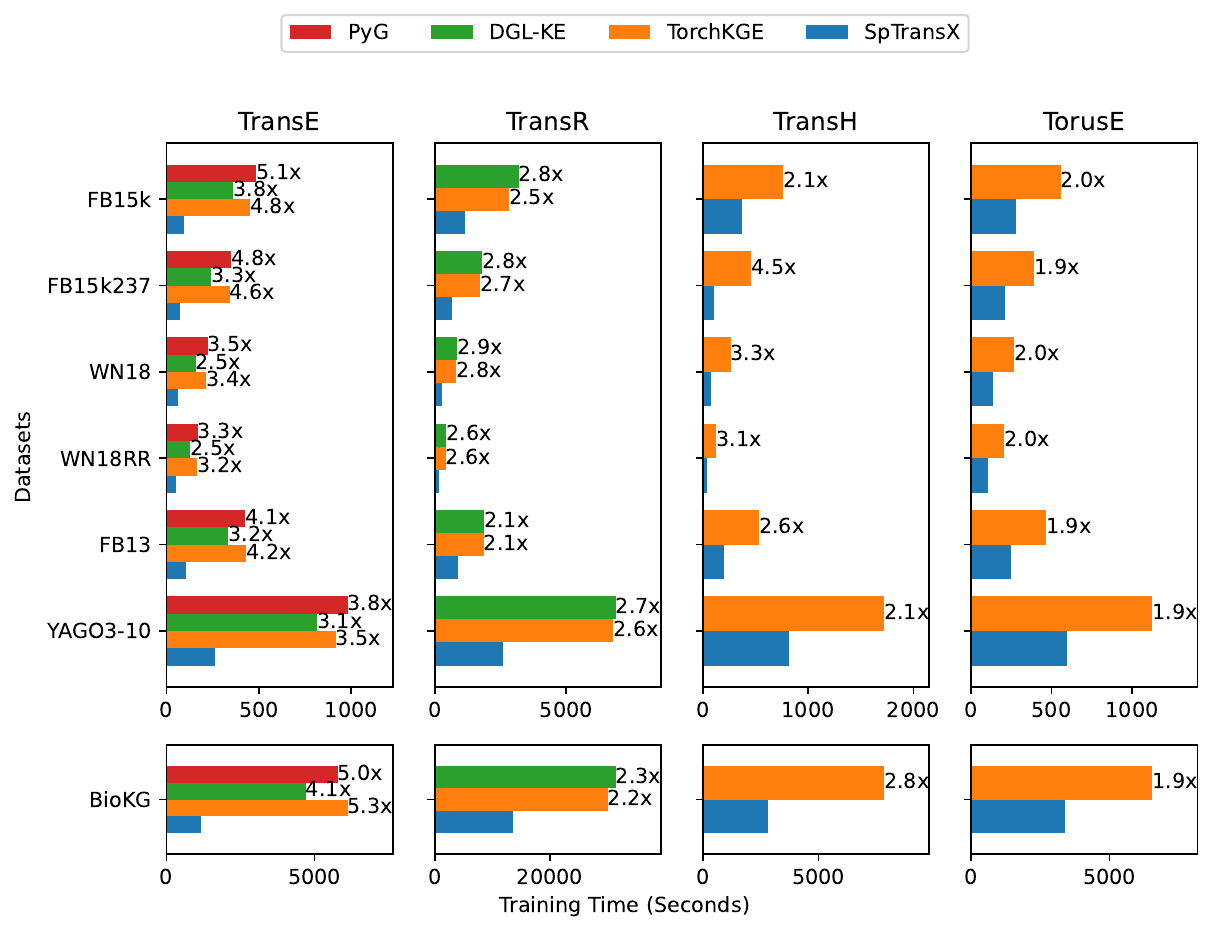}}
\subfigure[GPU]{\label{fig:b}\includegraphics[width=140mm]{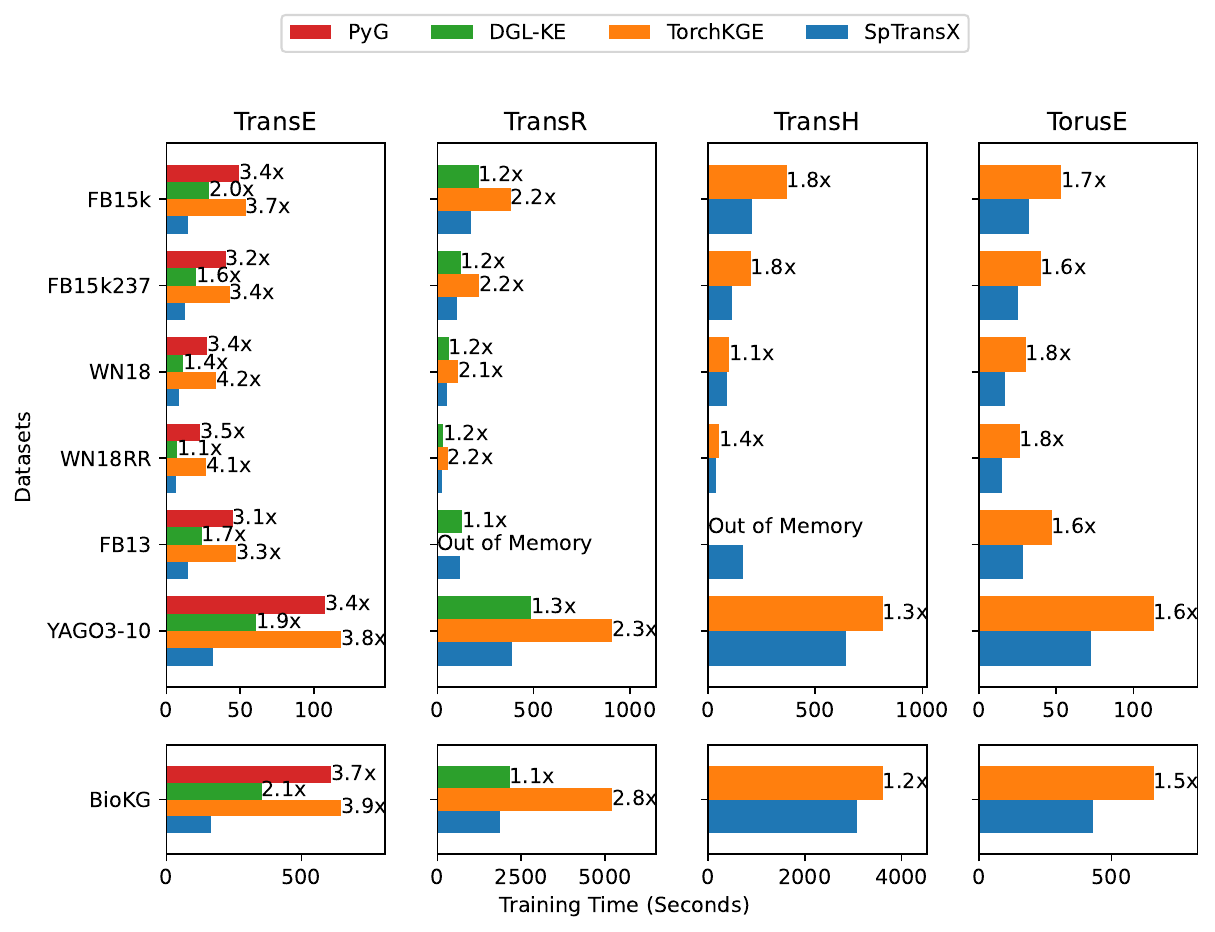}}
\caption{Total training time for CPU and GPU for various datasets. \revise{The slowdown factors of each framework compared to SpTransX are shown along the bars.}}
\label{fig:total_trg_time}
\end{figure*}

    The total training time for various datasets on CPU and GPU are shown in Figure \ref{fig:total_trg_time}. Our implementation outperforms all frameworks for both CPU and GPU. The speedup is consistent for both small and large datasets.

    SpTransX models exhibit good speedup on CPU and GPU systems. The speedups are consistent across datasets for the same model. We observe the most speedup in the TransE model. This is because, for this model, the computational bottleneck is the embedding gradient computation (see Figure \ref{fig:bottlenecks}). We eliminate this bottleneck by replacing fine-grained embedding scatter-gather with SpMM, which results in faster training time and efficient GPU memory usage (due to lower intermediate variable usage).
    
    Although TorusE uses the same scoring function, we do not observe the same amount of speedup in this model compared to TransE. This is because the primary computational bottleneck in this model is not always the embedding computation but the torus L2 dissimilarity function (marked as yellow boxes in Figure \ref{fig:bottlenecks}). 
    
    Among TransR and TransH, TransR is computationally more demanding. However, we still manage to perform better in TransR compared to TransH because the computational graph of TransH is much larger than TransR, and the embedding computation (or SpMM in our case) accounts for a lower percentage of system time compared to TransR. This means SpTransX has less impact on TransH compared to TransR.

\begin{figure*}[t]
\centering     
\subfigure[CPU]{\label{fig:cpu-break}\includegraphics[width=80mm]{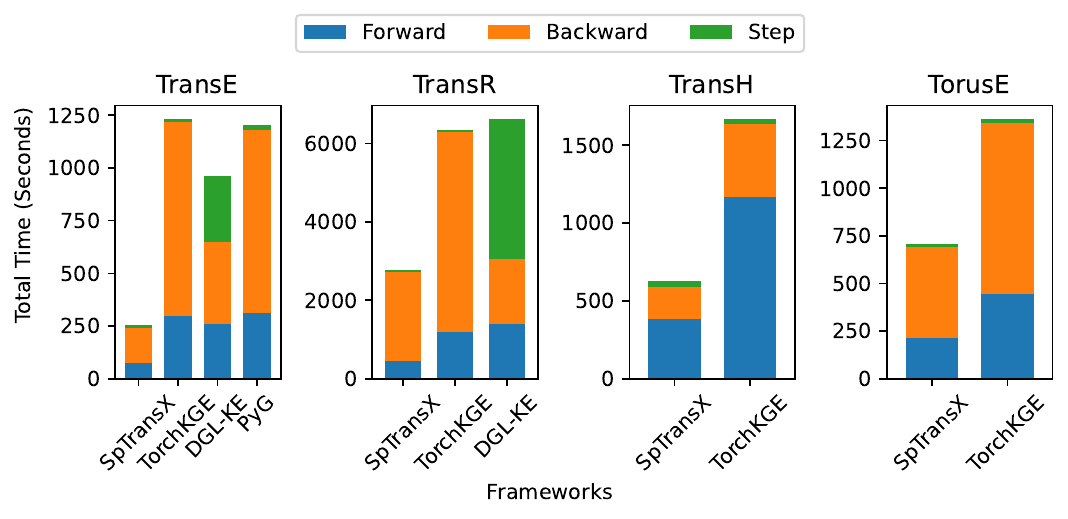}}
\hspace{5pt}
\subfigure[GPU]{\label{fig:gpu-break}\includegraphics[width=80mm]{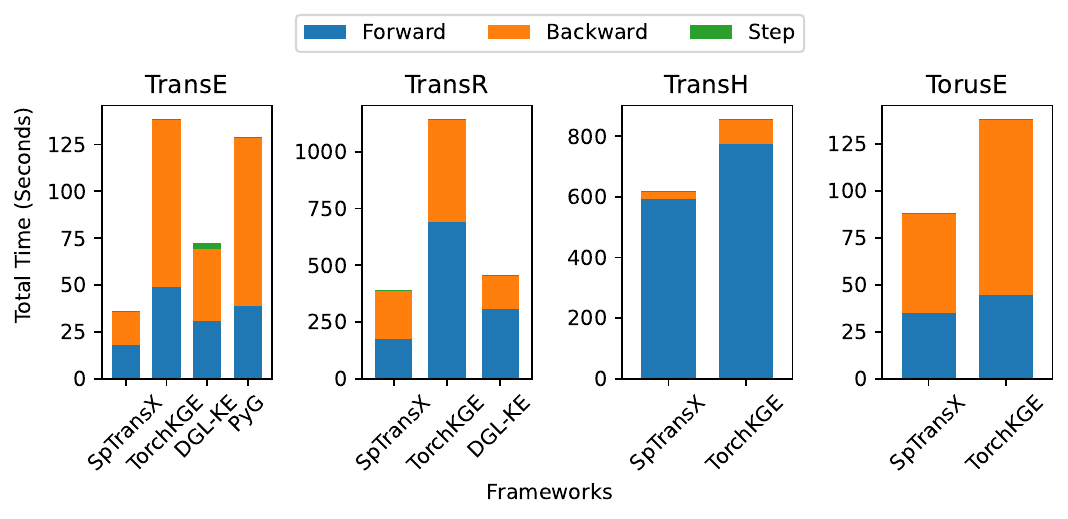}}
\caption{Breakdown of total training time for CPU and GPU on average of 7 datasets}
\label{fig:trg_breakdown}
\end{figure*}

    \subsubsection{GPU Memory Usage}
    Our implementations of the models take up significantly less CUDA memory than other frameworks. Table \ref{table:cuda-mem-allocation} demonstrates the average CUDA memory allocation for various frameworks and our implementations.

    \begin{table}[h]
    \caption{Average CUDA memory allocation for various models (in GB)}
    \label{table:cuda-mem-allocation}
    \vskip 0.15in
    \begin{center}
    \begin{small}
    \begin{sc}
    \begin{tabular}{ccccc}
    \toprule
    Model  & SpTransX       & TorchKGE & DGL-KE & PyG   \\
    \midrule
    TransE & \textbf{5.61}  & 13.55    & 11.37  & 13.54 \\
    TransR & \textbf{13.65} & 20.42    & 30.73  & -     \\
    TransH & \textbf{0.28}  & 3.1      & -      & -     \\
    TorusE & \textbf{12.03} & 15.87    & -      & -                   \\                        
    \bottomrule
    \end{tabular}
    \end{sc}
    \end{small}
    \end{center}
    \vskip -0.1in
    \end{table}

    SpTransX is optimized for GPU memory usage by limiting the model size to tensors only necessary during the training time. Furthermore, the SpMM accounts for fewer intermediate variables, reducing the memory footprint. We observe the highest GPU memory efficiency in the TransH model, around $11\times$ more efficient than TorchKGE on average. This is because the training loop uses linear algebraic implementation (discussed in subsection \ref{transh_formulation}) and reuses several expressions to reduce unnecessary GPU memory allocation.

\subsubsection{Breakdown of Training Time}
Each model training epoch consists of loss calculation (forward propagation), gradient computation (backward call), and parameter update (optimizer step). The chart in Figure \ref{fig:trg_breakdown} shows the average breakdown of the three steps for the frameworks.

We observe that SparsTransX improves the average forward propagation time for both CPU and GPU. It also outperforms backward computation for all cases except in TransR with DGL-KE for both GPU and CPU. DGL-KE uses the heterograph data structure instead of a regular triplet array and updates the backward gradients manually through the DGL graph API. This results in an unusually long parameter update time for DGL-KE in the CPU. This issue is not present in GPU since DGL-KE has a separate GPU implementation. Despite the slower backward time, SpTransX outperforms DGL-KE in terms of overall training time.

\begin{table}[h]
\centering
\caption{Average FLOPs count of 7 datasets \revise{(factor of $\times 10^{10}$)}}
\label{table:flops}
\begin{center}
\begin{small}
\begin{sc}
\begin{tabular}{ccccc}
\toprule
Model  & SpTransX       & TorchKGE & DGL-KE & PyG   \\
\midrule
TransE & \textbf{220}    & 483.87   & 293.06 & 483.82 \\
TransR & \textbf{567.37} & 1157.94  & 874.67 & -      \\
TransH & \textbf{9.66}   & 19.58    & -      & -      \\
TorusE & \textbf{289.99} & 387.93   & -      & -                        \\                        
\bottomrule
\end{tabular}
\end{sc}
\end{small}
\end{center}
\vskip -0.1in
\end{table}
\subsubsection{FLOPs count and Cache Miss Rate}
    We measure the FLOPs count for our CPU implementation and the cache miss rate. SpTransX exhibits a lower FLOP count than other frameworks for all models on average, as shown in Table \ref{table:flops}. It uses high-performance SpMM that typically uses fewer floating-point operations than regular non-sparse implementations. This results in the lowest average FLOP count for SpTransX compared to all frameworks for all models.

\begin{table}[h]
\caption{Average cache miss rate of 7 datasets \revise{(in \%)}}
\label{table:cache-miss}
\begin{center}
\begin{small}
\begin{sc}
\begin{tabular}{ccccc}
\toprule
Model  & SpTransX       & TorchKGE & DGL-KE & PyG   \\
\midrule
TransE    & \textbf{26.54} & 29.37         & 29.99  & 29.04 \\
TransR    & \textbf{17.02} & 19.20         & 29.54  & -     \\
TransH    & 10.43          & \textbf{9.75} & -      & -     \\
TorusE    & \textbf{21.53} & 22.94         & -      & -                 \\                        
\bottomrule
\end{tabular}
\end{sc}
\end{small}
\end{center}
\vskip -0.1in
\end{table}

Table \ref{table:cache-miss} lists the average cache miss rates. We observe that SpTransX performs better in all cases except for the TransH model. In this case, SparseTransX has a slightly higher cache miss rate than its peer, TorchKGE. This is because the impact of SpMM is small in the TransH model, and other operations overshadow the improved cache miss rate obtained by the efficient SpMM.


\subsubsection{Model Accuracies}
The sparse approach does not change the computational steps and thus does not affect the model accuracy. The accuracies of our implementations are consistent with that of other models, such as TorchKGE. For 100 epochs training on WN18 datasets \revise{with a fixed learning rate of 0.0004}, SpTransX's TransE, TorusE, and TransH models receive 0.72, 0.63, and 0.59 Hits@10 scores, whereas TorchKG's models receive 0.74, 0.63, and 0.60. \revise{A more detailed evaluation (discussed in Appendix \ref{A:eval}) reveals that SpTransX achieves similar or better Hits@10 accuracy compared to TorchKGE when the training loop is equipped with a learning rate scheduler.}

%% file: 7_section_discussion.tex



%% file: 8_section_conclusion.tex
\section{Conclusion}

Despite the inherent sparsity of knowledge graphs and their embedding algorithms, existing frameworks often do not leverage sparse matrix operations to accelerate the training of KGE models. 
We develop sparse formulations of translation-based KGE models that significantly outperform established knowledge graph frameworks, such as TorchKGE and DGL-KE, particularly regarding training time and GPU memory usage. Our findings demonstrate that the proposed approach consistently achieves improved performance across a range of both small and large datasets. \revise{We design a PyTorch-based library named SparseTransX that incorporates the sparse formulation methods for Knowledge Graph models and can demonstrate the aforementioned performance gains. The design of the library is flexible, and it can potentially perform distributed training once coupled with PyTorch Distributed Data Parallel (DDP) and Fully Sharded Data Parallel (FSDP) wrappers.}

Our sparse approach has multiple benefits. By using sparse representations, we reduce memory usage during training, which allows us to work with larger knowledge graphs without exhausting GPU resources. The efficiency improvements in training time come from optimizing matrix operations.
Given the extensive research in parallel sparse matrix operations and the availability of highly optimized libraries, our approach paves the way for faster computations and enhanced scalability for larger knowledge graphs. We believe this work will inspire further advancements in the development of robust and scalable knowledge graph frameworks. 

%% file: 9_appendix.tex
\newpage
\appendix
\section{Artifact Evaluation}
\label{A:AE}
\subsection{Abstract}


We provide a guideline to reproduce the training times for one of the datasets reported in this paper. We share a publicly available GitHub repository (\href{https://github.com/OnixHoque/sptransx-mlsys2025-reproduce}{https://github.com/OnixHoque/sptransx-mlsys2025-reproduce}) containing two bash scripts and a Jupyter Notebook that generate the workflow. The first bash script installs the frameworks needed to compare the performance with SpTransX and generates the environments required to run them. Another bash script runs the actual CPU and GPU experiments and stores the training times into text files. We include a validation Jupyter Notebook that performs processing on these text files and generates tables that depict Figure \ref{fig:total_trg_time} of the paper (for FB15K dataset). The experiments are run on one of the seven datasets (FB15K). The training time is computed for a single minibatch to keep the workflow short. The validation notebook approximates the total training time by multiplying the number of batches with the calculated training time. We further discuss how to run the workflow and perform validation from the generated results (also available in the Readme file of the GitHub repository). Please note that SpTransX is referred to as FastKG (former name) in the workflow.

\subsection{Artifact check-list (meta-information)}


{\small
\begin{itemize}
  \item {\bf Algorithm: } TransE, TransR, TransH, TorusE (Knowledge Graph Embedding Training Models)
  \item {\bf Program: } PyTorch
  \item {\bf Compilation: } Conda, PIP, Python 3.7 and Python 3.8
  \item {\bf Data set: } FB15K Knowledge Graph Dataset
  \item {\bf Run-time environment: } Conda, GCC
  \item {\bf Hardware: } A CPU (preferably AMD) with 64 cores and 512GB DDR4 memory. An NVIDIA GPU with 40GB VRAM.
  \item {\bf Run-time state: } Training Time
  \item {\bf Execution: } Bash Script and Jupyter Notebook (requires CUDA for GPU experiments)
  \item {\bf Metrics: } Training Time (in seconds)
  \item {\bf Output: } Table showing total training time and slowdown factors of other frameworks
  \item {\bf Experiments: } Performs a single minibatch training on FB15K dataset for 200 epochs for all models available in all four frameworks in CPU and GPU
  \item {\bf How much disk space required (approximately)?: } 50 GB
  \item {\bf How much time is needed to prepare workflow (approximately)?: } 3 hours
  \item {\bf How much time is needed to complete experiments (approximately)?: } 3 hours
  \item {\bf Publicly available?: } Yes
  \item {\bf Code licenses (if publicly available)?: } MIT
  \item {\bf Workflow framework used?: } Bash Scripts and Jupyter Notebook
  \item {\bf Archived (provide DOI)?: } Not generated yet
\end{itemize}

\subsection{Description}

\subsubsection{How delivered}

The workflow is available in the following GitHub repository: \href{https://github.com/OnixHoque/sptransx-mlsys2025-reproduce}{https://github.com/OnixHoque/sptransx-mlsys2025-reproduce}.

\subsubsection{Hardware dependencies}
The CPU and GPU experiments were run on dedicated CPU/GPU (single) nodes of NERSC Perlmutter. Their configurations are given below. The parameters are set to maximize CPU/GPU utilization. Similar configurations are recommended to reproduce the results. The CPU configuration is AMD EPYC 7763 (Milan) CPU with 64 cores and 512GB DDR4 memory. The GPU configuration is a single NVIDIA A100-SXM4 GPU with 40 GB VRAM.

\subsubsection{Software dependencies}
The experiments were tested on the following configuration.
\begin{itemize}
    \item GCC 12.2
    \item Conda 24.9.1
    \item Python 3.9 (3.8 for DGLKE)
    \item PyTorch 2.3.1 (1.7.1 for DGLKE)
    \item CUDAToolKit 12.1 (11.0 for DGLKE)
\end{itemize}

\subsubsection{Datasets}
FB15K dataset is used in the workflow. It is one of the seven datasets used in the paper. It is included in the repository.
\subsection{Environment Installation}
To set up the environments, clone the GitHub repository and run the following command. It will create two virtual environments. One specific to DGLKE, and another for the rest.

\begin{verbatim}
    ./0.setup_environments.sh
\end{verbatim}

\subsection{Experiment Workflow}
To run the experiments, execute the following command. 
\begin{verbatim}
    ./1.run_experiments.sh
\end{verbatim}
It will generate the training time of a single minibatch training for various models and frameworks of the FB15K dataset. The outputs will be saved in cpu.txt and gpu.txt.
\subsection{Evaluation and Expected Result}
To generate the table of Figure \ref{fig:total_trg_time} (for FB15K) in the paper, execute the Jupyter Notebook 2.validation.ipynb. It will parse the generated text files and produce the tables for CPU and GPU for the FB15k dataset. The table includes the total training time and the slowdown factors of each framework compared to SpTransX. 

It is expected that SpTransX is up to $5\times$ and $4\times$ faster in CPU and GPU, respectively, compared to other frameworks. The Jupyter Notebook contains the expected tables for both CPU and GPU from past experiments.


\subsection{Notes}
Please note that \textbf{SpTransX} is referred to as \textbf{FastKG} (former name) in the workflow.



\section{Applicability of SparseTransX for dense graphs} 
\label{A:density}
Even for fully dense graphs, our KGE computations remain highly sparse. This is because our SpMM leverages the incidence matrix for triplets, rather than the graph's adjacency matrix. In the paper, the sparse matrix $A \in \{-1,0,1\}^{M \times (N+R)}$ represents the triplets, where $N$ is the number of entities, $R$ is the number of relations, and $M$ is the number of triplets. This representation remains extremely sparse, as each row contains exactly three non-zero values (or two in the case of the "ht" representation). Hence, the sparsity of this formulation is independent of the graph's structure, ensuring computational efficiency even for dense graphs.

\section{Computational Complexity}
\label{A:complexity}
 For a sparse matrix $A$ with $m \times k$ having $nnz(A)=$ number of non zeros and dense matrix $X$ with $k \times n$ dimension, the computational complexity of the SpMM is $O(nnz(A) \cdot n)$ since there are a total of $nnz(A)$ number of dot products each involving $n$ components. Since our sparse matrix contains exactly three non-zeros in each row, $nnz(A) = 3m$. Therefore, the complexity of SpMM is $O(3m \cdot n)$ or $O(m \cdot n)$, meaning the complexity increases when triplet counts or embedding dimension is increased. Memory access pattern will change when the number of entities is increased and it will affect the runtime, but the algorithmic complexity will not be affected by the number of entities/relations.

\section{Applicability to Non-translational Models}
\label{A:non_trans}
Our paper focused on translational models using sparse operations, but the concept extends broadly to various other knowledge graph embedding (KGE) methods. Neural network-based models, which are inherently matrix-multiplication-based, can be seamlessly integrated into this framework. Additionally, models such as DistMult, ComplEx, and RotatE can be implemented with simple modifications to the SpMM operations. Implementing these KGE models requires modifying the addition and multiplication operators in SpMM, effectively changing the semiring that governs the multiplication.   

In the paper, the sparse matrix $A \in \{-1,0,1\}^{M \times (N+R)}$ represents the triplets, and the dense matrix $E \in \mathbb{R}^{(N+R) \times d}$ represents the embedding matrix, where $N$ is the number of entities, $R$ is the number of relations, and $M$ is the number of triplets. TransE’s score function, defined as $h + r - t$, is computed by multiplying $A$ and $E$ using an SpMM followed by the L2 norm. This operation can be generalized using a semiring-based SpMM model: $Z_{ij} = \bigoplus_{k=1}^{n} (A_{ik} \otimes E_{kj})$

Here, $\oplus$ represents the semiring addition operator, and $\otimes$ represents the semiring multiplication operator. For TransE, these operators correspond to standard arithmetic addition and multiplication, respectively.

\subsection*{DistMult} 
DistMult’s score function has the expression $h \odot r \odot t$. To adapt SpMM for this model, two key adjustments are required: The sparse matrix $A$ stores $+1$ at the positions corresponding to $h_{\text{idx}}$, $t_{\text{idx}}$, and $r_{\text{idx}}$. Both the semiring addition and multiplication operators are set to arithmetic multiplication. These changes enable the use of SpMM for the DistMult score function.

\subsection*{ComplEx} 
ComplEx’s score function has $h \odot r \odot \bar{t}$, where embeddings are stored as complex numbers (e.g., using PyTorch). In this case, the semiring operations are similar to DistMult, but with complex number multiplication replacing real number multiplication.

\subsection*{RotatE} 
RotatE’s score function has $h \odot r - t$. For this model, the semiring requires both arithmetic multiplication and subtraction for $\oplus$. With minor modifications to our SpMM implementation, the semiring addition operator can be adapted to compute $h \odot r - t$.

\subsection*{Support from other libraries}
Many existing libraries, such as GraphBLAS (Kimmerer, Raye, et al., 2024), Ginkgo (Anzt, Hartwig, et al., 2022), and Gunrock (Wang, Yangzihao, et al., 2017), already support custom semirings in SpMM. We can leverage C++ templates to extend support for KGE models with minimal effort.

\begin{figure*}[t]
\centering     
\includegraphics[width=\textwidth]{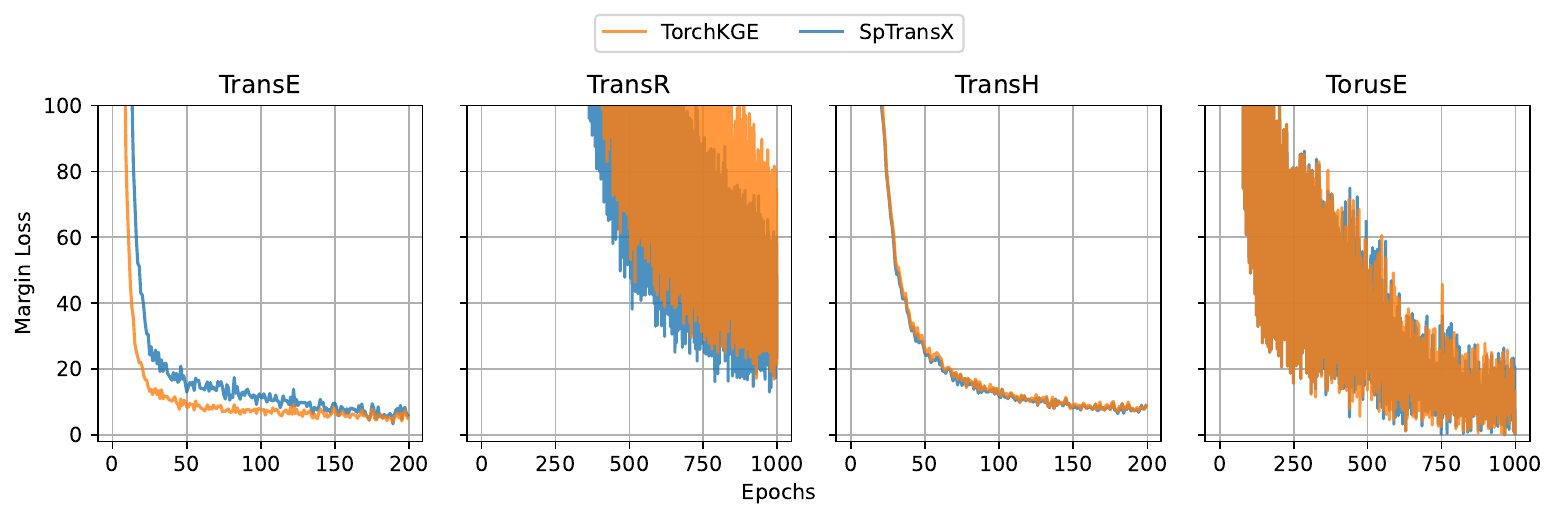}
\caption{Loss curve for sparse and non-sparse approach. The sparse approach eventually reaches the same loss value with similar Hits@10 test accuracy.}
\label{fig:loss_curve}
\end{figure*}

\begin{table}[h]
\centering
\caption{Average of 9 Hits@10 Accuracy for WN18 dataset}
\begin{tabular}{|c|c|c|}
\hline
\textbf{Model} & \textbf{TorchKGE} & \textbf{SpTransX} \\ \hline
TransE         & 0.79 ± 0.001700   & 0.79 ± 0.002667   \\ \hline
TransR         & 0.29 ± 0.005735   & 0.33 ± 0.006154   \\ \hline
TransH         & 0.76 ± 0.012285   & 0.79 ± 0.001832   \\ \hline
TorusE         & 0.73 ± 0.003258   & 0.73 ± 0.002780   \\ \hline
\end{tabular}
\label{table:perf_eval}
\end{table}

\section{Model Performance Evaluation and Convergence}
\label{A:eval}
SpTransX follows a slightly different loss curve (see Figure \ref{fig:loss_curve}) and eventually converges with the same loss as other non-sparse implementations such as TorchKGE. We test SpTransX with the WN18 dataset having embedding size 512 (128 for TransR and TransH due to memory limitation) and run 200-1000 epochs. We compute average Hits@10 of 9 runs with different initial seeds and a learning rate scheduler. The results are shown below. We find that Hits@10 is generally comparable to or better than the Hits@10 achieved by TorchKGE.

\section{Preliminarily Experiment on Scaling}
\label{A:scaling}
Communication can be a significant bottleneck in distributed KGE training when using sparse-dense matrix multiplication (SpMM) kernels. We perform a preliminary experiment to observe the scaling capacity of SparseTransX for the TransE model using a large knowledge graph dataset, COVID-19 \cite{tabassum2024knowledge}. We wrap the model in the Data Distributed Parallel (DDP) wrapper provided by PyTorch and train the model to 64 NVIDIA A100 GPUs to observe scaling behaviour. DDP (Distributed Data Parallel) trains a model by replicating it across multiple GPUs, synchronizing gradients during backpropagation to ensure consistent updates. The training times for various GPUs are reported in Table \ref{table:scaling}.

\begin{table}[h]
\centering
\caption{Scaling TransE model on COVID-19 dataset with 60,820 entities, 62 relations, and 1,032,939 triplets \cite{tabassum2024knowledge}}
\begin{tabular}{|c|c|}
\hline
\textbf{Number of GPUs} & \textbf{500 epoch time (seconds)} \\ \hline
4                       & 706.38                            \\ \hline
8                       & 586.03                            \\ \hline
16                      & 340.00                               \\ \hline
32                      & 246.02                            \\ \hline
64                      & 179.95                            \\ \hline
\end{tabular}
\label{table:scaling}
\end{table}

It indicates that communication is not a bottleneck up to 64 GPUs. If communication becomes a performance bottleneck at larger scales, we plan to explore alternative communication-reducing algorithms, including 2D and 3D matrix distribution techniques, which are known to minimize communication overhead at extreme scales. Additionally, we will incorporate model parallelism alongside data parallelism for large-scale knowledge graphs.

\section{Backpropagation of SpMM}
\label{A:backprop}
 Our main computational kernel is the sparse-dense matrix multiplication (SpMM). The computation of backpropagation of an SpMM w.r.t. the dense matrix is also another SpMM. To see how, let's consider the sparse-dense matrix multiplication $AX = C$ which is part of the training process. As long as the computational graph reduces to a single scaler loss $\mathfrak{L}$, it can be shown that $\frac{\partial C}{\partial X} = A^T$. Here, $X$ is the learnable parameter (embeddings), and $A$ is the sparse matrix. Since $A^T$ is also a sparse matrix and $\frac{\partial \mathfrak{L}}{\partial C}$ is a dense matrix, the computation $\frac{\partial \mathfrak{L}}{\partial X} = \frac{\partial C}{\partial X} \times \frac{\partial \mathfrak{L}}{\partial C} = A^T \times \frac{\partial \mathfrak{L}}{\partial C} $ is an SpMM. This means that both forward and backward propagation of our approach benefit from the efficiency of a high-performance SpMM.

\subsection*{Proof that $\frac{\partial C}{\partial X} = A^T$}
 To see why $\frac{\partial C}{\partial X} = A^T$ is used in the gradient calculation, we can consider the following small matrix multiplication without loss of generality.
\begin{align*}
A &= \begin{bmatrix}
a_1 & a_2 \\
a_3 & a_4
\end{bmatrix} \\ 
 X &= \begin{bmatrix}
x_1 & x_2 \\
x_3 & x_4
\end{bmatrix} \\
 C &=  \begin{bmatrix}
c_1 & c_2 \\
c_3 & c_4
\end{bmatrix}
\end{align*}
Where $C=AX$, thus-
\begin{align*}
c_1&=f(x_1, x_3) \\
c_2&=f(x_2, x_4) \\
c_3&=f(x_1, x_3) \\
c_4&=f(x_2, x_4) \\
\end{align*}
Therefore-
\begin{align*}
\frac{\partial \mathfrak{L}}{\partial x_1} &= \frac{\partial \mathfrak{L}}{\partial c_1} \times \frac{\partial c_1}{\partial x_1} + \frac{\partial \mathfrak{L}}{\partial c_2} \times \frac{\partial c_2}{\partial x_1} + \frac{\partial \mathfrak{L}}{\partial c_3} \times \frac{\partial c_3}{\partial x_1} + \frac{\partial \mathfrak{L}}{\partial c_4} \times \frac{\partial c_4}{\partial x_1}\\
&= \frac{\partial \mathfrak{L}}{\partial c_1} \times \frac{\partial \mathfrak{c_1}}{\partial x_1} + 0 + \frac{\partial \mathfrak{L}}{\partial c_3} \times \frac{\partial \mathfrak{c_3}}{\partial x_1} + 0\\
&= a_1 \times \frac{\partial \mathfrak{L}}{\partial c_1} + a_3 \times \frac{\partial \mathfrak{L}}{\partial c_3}\\
\end{align*}

Similarly-
\begin{align*}
\frac{\partial \mathfrak{L}}{\partial x_2}
&= a_1 \times \frac{\partial \mathfrak{L}}{\partial c_2} + a_3 \times \frac{\partial \mathfrak{L}}{\partial c_4}\\
\frac{\partial \mathfrak{L}}{\partial x_3}
&= a_2 \times \frac{\partial \mathfrak{L}}{\partial c_1} + a_4 \times \frac{\partial \mathfrak{L}}{\partial c_3}\\
\frac{\partial \mathfrak{L}}{\partial x_4}
&= a_2 \times \frac{\partial \mathfrak{L}}{\partial c_2} + a_4 \times \frac{\partial \mathfrak{L}}{\partial c_4}\\
\end{align*}
This can be expressed as a matrix equation in the following manner-
\begin{align*}
\frac{\partial \mathfrak{L}}{\partial X} &= \frac{\partial C}{\partial X} \times \frac{\partial \mathfrak{L}}{\partial C}\\
\implies \begin{bmatrix}
\frac{\partial \mathfrak{L}}{\partial x_1} & \frac{\partial \mathfrak{L}}{\partial x_2} \\
\frac{\partial \mathfrak{L}}{\partial x_3} & \frac{\partial \mathfrak{L}}{\partial x_4}
\end{bmatrix} &= \frac{\partial C}{\partial X} \times \begin{bmatrix}
\frac{\partial \mathfrak{L}}{\partial c_1} & \frac{\partial \mathfrak{L}}{\partial c_2} \\
\frac{\partial \mathfrak{L}}{\partial c_3} & \frac{\partial \mathfrak{L}}{\partial c_4}
\end{bmatrix}
\end{align*}
By comparing the individual partial derivatives computed earlier, we can say-

\begin{align*}
\begin{bmatrix}
\frac{\partial \mathfrak{L}}{\partial x_1} & \frac{\partial \mathfrak{L}}{\partial x_2} \\
\frac{\partial \mathfrak{L}}{\partial x_3} & \frac{\partial \mathfrak{L}}{\partial x_4}
\end{bmatrix} &= \begin{bmatrix}
a_1 & a_3 \\
a_2 & a_4
\end{bmatrix} \times \begin{bmatrix}
\frac{\partial \mathfrak{L}}{\partial c_1} & \frac{\partial \mathfrak{L}}{\partial c_2} \\
\frac{\partial \mathfrak{L}}{\partial c_3} & \frac{\partial \mathfrak{L}}{\partial c_4}
\end{bmatrix}\\
\implies \begin{bmatrix}
\frac{\partial \mathfrak{L}}{\partial x_1} & \frac{\partial \mathfrak{L}}{\partial x_2} \\
\frac{\partial \mathfrak{L}}{\partial x_3} & \frac{\partial \mathfrak{L}}{\partial x_4}
\end{bmatrix} &= A^T \times \begin{bmatrix}
\frac{\partial \mathfrak{L}}{\partial c_1} & \frac{\partial \mathfrak{L}}{\partial c_2} \\
\frac{\partial \mathfrak{L}}{\partial c_3} & \frac{\partial \mathfrak{L}}{\partial c_4}
\end{bmatrix}\\
\implies \frac{\partial \mathfrak{L}}{\partial X} &= A^T \times \frac{\partial \mathfrak{L}}{\partial C}\\
\therefore \frac{\partial C}{\partial X} &= A^T \qed
\end{align*}